%% file: arxiv.tex
\documentclass[letterpaper]{article} 
\usepackage[submission]{aaai25}  

\usepackage{times}  
\usepackage{helvet}  
\usepackage{courier}  
\usepackage[hyphens]{url}  
\usepackage{graphicx} 
\usepackage{natbib}  
\usepackage{hyperref}

\usepackage{caption} 
\usepackage{algorithm}
\usepackage{amsmath}
\usepackage[capitalize]{cleveref}
\crefname{section}{Sec.}{Secs.}
\Crefname{section}{Section}{Sections}
\Crefname{table}{Table}{Tables}
\crefname{table}{Tab.}{Tabs.}
\usepackage{comment}

\usepackage{xcolor,pifont}
\newcommand*\colourcheck[1]{%
  \expandafter\newcommand\csname #1check\endcsname{\textcolor{#1}{\ding{52}}}%
}
\newcommand*\colourcross[1]{%
  \expandafter\newcommand\csname #1check\endcsname{\textcolor{#1}{\ding{55}}}%
}
\definecolor{my_green}{RGB}{51,102,0}
\definecolor{my_red}{RGB}{204, 0, 0}

\newcommand{\cmark}{\textcolor{my_green}{\ding{51}}} 
\newcommand{\xmark}{\textcolor{my_red}{\ding{55}}} 
\colourcheck{green}
\colourcross{red}
\usepackage{bbding}
\usepackage{algorithm}
\usepackage{array}
\usepackage{amsfonts}
\usepackage{multirow}
\usepackage{amsmath}
\usepackage{mathalfa}
\usepackage{makecell}
\usepackage{breqn}
\usepackage[bottom]{footmisc}

\usepackage{xcolor}
\usepackage{listings}
\usepackage{color}
\usepackage{adjustbox}
\usepackage{}
\usepackage{multirow}
\usepackage{booktabs}
\usepackage{bbding}
\usepackage{adjustbox}
\usepackage{caption}
\usepackage{tabularx}
\usepackage{cleveref}
\usepackage{xcolor,pifont}
\definecolor{mygray-bg}{gray}{0.9}
\usepackage{tcolorbox}
\usepackage{colortbl}
\usepackage{float}

\usepackage{paralist}

\usepackage[utf8]{inputenc} 
\usepackage[T1]{fontenc}    

\usepackage{url}            
\usepackage{booktabs}       
\usepackage{amsfonts}       
\usepackage{nicefrac}       
\usepackage{microtype}      
\usepackage{xcolor}       
\usepackage{cleveref}
\usepackage{subcaption}
\usepackage{graphicx}
\usepackage{booktabs}
\usepackage{enumerate}
\usepackage{xcolor}
\usepackage{multicol}
\usepackage{multirow}
\usepackage{enumitem}
\usepackage{colortbl}
\usepackage{lipsum,booktabs}
\usepackage{tcolorbox}
\usepackage{dashrule}
\usepackage{xspace}
\usepackage{algpseudocode}

\newcommand{\vref}{\mathbf{v}}
\newcommand{\vhyp}{\mathbf{\hat{v}}}
\newcommand{\senthyp}{\hat{v}}
\newcommand{\txthyp}{\hat{t}}
\newcommand{\methodp}{P_{CLIP}}
\newcommand{\methodr}{R_{CLIP}}
\newcommand{\methodf}{F_{CLIP}}

\usepackage{newfloat}
\usepackage{listings}
\DeclareCaptionStyle{ruled}{labelfont=normalfont,labelsep=colon,strut=off} 
\floatstyle{ruled}
\newfloat{listing}{tb}{lst}{}
\floatname{listing}{Listing}
\definecolor{lightpurple}{RGB}{230, 190, 255}

\title{V2Xum-LLM: Cross-Modal Video Summarization with\\Temporal Prompt Instruction Tuning}
\author{
    Hang Hua\equalcontrib,
    Yunlong Tang\equalcontrib,
    Chenliang Xu,
    Jiebo Luo\thanks{Corresponding author.}
}
\affiliations{
    University of Rochester
     \\
  {\tt \small
    \{hhua2,jluo\}@cs.rochester.edu
  }\\
  {\tt \small
   \{yunlong.tang,chenliang.xu\}@rochester.edu
  }\\
   {\tt
   \href{https://hanghuacs.github.io/v2xum/}{\url{https://hanghuacs.github.io/v2xum/}}
   }
}

\usepackage{bibentry}

\begin{document}

\maketitle

\input{sec/0_abstract}
\input{sec/1_intro}

\input{sec/2_related}

\input{sec/3_benchmark}

\input{sec/4_method}
\input{sec/5_experiments}
\input{sec/6_revisiting}

\input{sec/7_conclusion}

\bibliography{aaai25}
\input{sec/8_appendix}

\end{document}

%% file: sec/0_abstract.tex
\begin{abstract}
Video summarization aims to create short, accurate, and cohesive summaries of longer videos. Despite the existence of various video summarization datasets, a notable limitation is their limited amount of source videos, which hampers the effective training of advanced large vision-language models (VLMs). Additionally, most existing datasets are created for video-to-video summarization, overlooking the contemporary need for multimodal video content summarization. Recent efforts have been made to expand from unimodal to multimodal video summarization, categorizing the task into three sub-tasks based on the summary's modality: video-to-video (V2V), video-to-text (V2T), and a combination of video and text summarization (V2VT). However, the textual summaries in previous multimodal datasets are inadequate. 
To address these issues, we introduce Instruct-V2Xum, a cross-modal video summarization dataset featuring 30,000 diverse videos sourced from YouTube, with lengths ranging from 40 to 940 seconds and an average summarization ratio of 16.39\%. Each video summary in Instruct-V2Xum is paired with a textual summary that references specific frame indexes, facilitating the generation of aligned video and textual summaries.
In addition, we propose a new video summarization framework named V2Xum-LLM. V2Xum-LLM, specifically V2Xum-LLaMA in this study, is the first framework that unifies different video summarization tasks into one large language model's (LLM) text decoder and achieves task-controllable video summarization with temporal prompts and task instructions. Experiments show that V2Xum-LLaMA outperforms strong baseline models on multiple video summarization tasks. 
Furthermore, we propose an enhanced evaluation metric for V2V and V2VT summarization tasks.
\end{abstract}

%% file: sec/1_intro.tex
\section{Introduction}
\label{sec:intro}

\begin{figure}
    \centering
    \includegraphics[width=0.95\columnwidth]{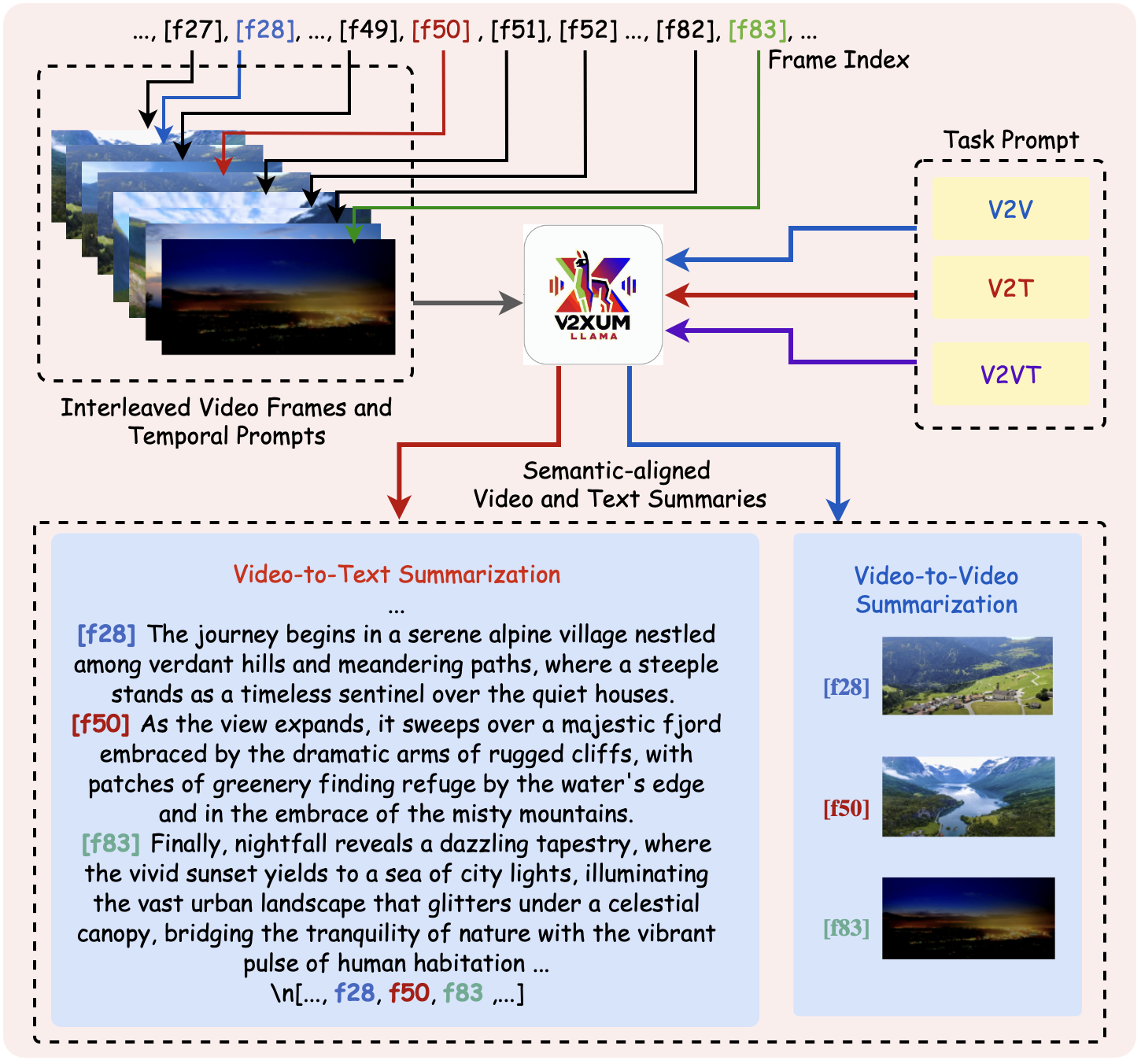}
    \caption{Illustration of cross-modal video summarization.}
   \vspace{-5mm}
    \label{fig:teaser}
\end{figure}

The interest in sharing life experiences has surged in recent years, making video the most informative and diverse visual medium on social media platforms. This trend has led to significant demands for a variety of video and language understanding tasks, such as video captioning \cite{xu2016msr}, video question answering \cite{yu2019activitynet,xiao2021next}, moment retrieval \cite{lei2021detecting}, and video summarization \cite{gygli2014creating,song2015tvsum}. Video summarization (V2V) provides an efficient way for humans to obtain key information from a long video. This process entails selecting the most significant information from a video and condensing it into a shorter form, all while maintaining the essence of the original content.  Beyond extensive research in V2V summarization, there are a few recent explorations of V2T and V2VT summarization. Most notably, VideoXum \cite{lin2023videoxum} seeks to broaden the modality of video summaries to include text summaries. It utilizes the dense captions from ActivityNetCap \cite{krishna2017dense} videos as text summaries and annotates the corresponding video segments as summaries. 
Other datasets for video summarization include TVSum \cite{song2015tvsum}, SumMe \cite{gygli2014creating}, QFVS \cite{sharghi2017query}, MED Summaries \cite{potapov2014category}, and so on. While it is expected that powerful LLMs can help improve video summarization, 
the insufficient number of source videos may not be able to support the robust fine-tuning of LLMs to perform this task since finetuning a large model with a limited number of training examples is prone to overfitting \cite{hua2021noise,hua2023improving}.
In addition, VideoXum data cannot be considered true summaries due to redundant information in ActivityNetCap's dense captions. More recently, Shot2Story20K \cite{han2023shot2story20k} collects 20k video-text data that enables the robust finetuning of LLMs, but it only supports video-to-text summarization. To address these issues, we propose Instruct-V2Xum, a new large-scale cross-model video summarization dataset that contains 30k open domain videos, partitioned as 25,000 in the training set, 1,000 in the validation set, and 4,000 in the test set. In Instruct-V2Xum, we obtain the source videos from YouTube using the video list provided by InternVid \cite{wang2023internvid}. The methodology of video summarization parallels that of extractive text summarization, where the objective is to isolate the pivotal frames or sentences from the source videos or documents, respectively. Extractive text summarization is a foundational task in the NLP field, with numerous LLMs setting the benchmark in this domain \cite{zhang2023diffusum,jia2020neural,a2summ}. Inspired by this, we first extract frames from the source videos and employ LLaVA-1.5-7B \cite{llava} to generate detailed captions for each frame. Then we take all the frame captions as a document to perform extractive document summarization using GPT-4 \cite{achiam2023gpt}. This approach enables us to obtain both video summaries and their corresponding textual summaries. Finally, the extracted text summaries are further refined by GPT-4.

Numerous visual instruction tuning-based methods have been proposed for video-language understanding \cite{huang2023vtimellm,wang2023chatvideo,video-llava}. These models can process the long videos for general video-language understanding and reasoning tasks such as video question answering, video captioning, and so on. Recently, there have been some attempts for fine-grained video moments understanding \cite{huang2023vtimellm} or video-language temporal grounding \cite{lin2023univtg} using large VLMs. However, these approaches, which typically process video frames as sequential images for a frozen visual encoder and train an LLM decoder for identifying video moment boundaries, are not well-suited for dense temporal prediction in video summarization tasks, particularly in the V2VT tasks. Furthermore, most existing models require large-scale data to train new parameters added to pre-trained VLMs \cite{lin2023univtg,video-llava,video-chatgpt,he2023a2summ}. This requirement significantly limits their practicality in scenarios where only a limited number of training examples are available. To address these problems, we design a new temporal prompt instruction tuning framework -- V2Xum-LLaMA. In V2Xum-LLaMA, we unify different modalities of video summary generation into one LLM decoder. This framework stands out by removing the dependence on task-specific layers that were required for video summarization in earlier VLM-based approaches. The main advantages of this method are that it enables the effective adaptation of the learned knowledge and the powerful capabilities of the pretrained language models to dense video temporal and content understanding. All the pretrained parameters of the VLMs are reused, and the model takes interleaved video frames and natural language temporal prompts as input to facilitate end-to-end model training. As video temporal prediction is performed by using language models, there is a challenge for calculating the correlation for V2V evaluation, since the language model-predicted video summaries are the discrete frame indexes. To overcome this challenge, we propose a solution to calculate the scores for language model-predicted video summaries. Furthermore, we also provide an analysis of the existing video summarization tasks and propose $F_{CLIP}$ and $Cross-F_{CLIP}$, the enhanced evaluation metrics for V2V and V2VT summarization tasks.

In summary, our main contributions are as follows:
\begin{itemize}
    \item We propose V2Xum-LLaMA, a novel cross-modal video summarization framework that unifies different tasks into a single pre-trained language decoder, eliminating the need for task-specific heads used in prior methods. By taking interleaved video frames and temporal prompts as input, our method enables end-to-end processing of long video sequences and outperforms all strong baseline models on mainstream V2V, V2T, and V2VT benchmarks.
    \item To address the lack of video-language data for fine-tuning large VLMs in video summarization tasks, we created Instruct-V2Xum, a new instruction-following dataset for cross-modal video summarization. It contains 30k diverse YouTube videos, ranging from 40 to 940 seconds, enabling VLMs to generate modality-controllable video summaries with task prompts. The experiments validate the rationality of our proposed dataset.
    \item We present a comprehensive analysis of the limitations in current video summarization tasks from the perspectives of data, methods, and evaluation. Based on this, we propose $F_{CLIP}$ and $Cross-F_{CLIP}$, an enhanced evaluation metric for V2V and V2VT summarization tasks. Experimental results show that these metrics are highly consistent with the traditional evaluation metrics including F1, Spearman correlation, and Kendall correlation.
\end{itemize}

%% file: sec/2_related.tex
\section{Related Work}
\input{table/data_compare}
\subsection{Video Summarization}

 Traditional video summarization, also known as video-to-video summarization, typically generates a condensed version of the original video, comprising selected frames~\cite{liu2020transforming, ghauri2021supervised, wang2019stacked, fajtl2019summarizing}, shots~\cite{ji2019video, feng2018extractive, zhang2018retrospective}, or segments~\cite{tang2022multi, koutras2019susinet}. These models are commonly trained using supervised learning approaches, with reinforcement learning methods like policy gradient~\cite{williams1992simple}, optimizing for diversity and representativeness in the summarized output~\cite{zhou2018dsn}, gaining popularity.
The standard datasets for video summarization tasks include SumMe~\cite{gygli2014creating} and TVSum~\cite{song2015tvsum}, which are widely used for benchmarking purposes. In recent years, cross-modal video summarization~\cite{fu2020multi, haopeng2022progressive, huang2021gpt2mvs} has emerged as an area of interest, incorporating additional modalities such as audio, speech, subtitles, and captions. These approaches leverage multimodal models to create more comprehensive summaries. Video-to-text summarization~\cite{palaskar2019multimodal,choi2018contextually} is an evolving field that aims to generate descriptive paragraphs in natural language that encapsulate video content. VideoXum~\cite{lin2023videoxum} advances this field by repurposing the ActivityNetCap~\cite{krishna2017dense} dataset and employing the BLIP-2~\cite{li2023blip} model for both V2V and V2T summarization. However, the data in~\cite{lin2023videoxum} may not be genuine summaries, as the dense captions from ActivityNetCap often include significant redundancy.

\subsection{Large Language Models}
In recent years, Large Language Models (LLMs) have witnessed rapid advancements~\cite{achiam2023gpt,touvron2023llama,touvron2023llama2,zhang2023llama-adapter}. With pretraining on extensive corpora from the Internet, LLMs acquire substantial knowledge, enabling powerful zero-shot and in-context learning capabilities~\cite{achiam2023gpt,wei2022emergent,kaplan2020scaling,hu2022promptcap}. Efforts have been increasingly directed toward leveraging LLMs for multimodal tasks~\cite{macawllm,avllm,pandagpt,yu2024promptfix,hua2024finematch}. Techniques such as vision-language alignment and adapter fine-tuning are employed to integrate LLMs into the multimodal domain. These methods align the visual features extracted by visual encoders with the input token space of LLMs~\cite{llava}. Typically, the parameters of LLMs are frozen to retain their existing capabilities although LoRA~\cite{lora} fine-tuning is sometimes applied in low-resource settings. Based on this, several studies~\cite{wang2023chatvideo,video-chatgpt, video-llava,videollama} have successfully employed LLMs for video understanding tasks, referred to as Vid-LLMs~\cite{tang2023video}. However, current research primarily concentrates on general video understanding tasks, such as video question-answering (QA) and video captioning, with less emphasis on temporal information. Recent works~\cite{huang2023vtimellm,ren2023timechat,tang2024avicuna} explore LLMs' potential in temporal grounding and localization, emphasizing their untapped capability in video temporal understanding.

%% file: table/data_compare.tex
\begin{table*}[htbp]
\centering
\resizebox{1\linewidth}{!}{%
\begin{tabular}{@{}l|llccccc@{}}
\toprule
\textbf{Dataset} & \textbf{Domain} & \textbf{\# Videos} & \textbf{Annotation} & \textbf{V2V-Sum} & \textbf{V2T-Sum} & \textbf{V2VT-Sum} & \textbf{Instruction} \\ \midrule
\midrule
MSVD~\cite{Chen2011CollectingHP} & Open & 1,970 & M & {\scriptsize \xmark} & \cmark & {\scriptsize \xmark} & {\scriptsize \xmark} \\
YouCook~\cite{das2013thousand} & Cooking & 88 & M & {\scriptsize \xmark} & \cmark & {\scriptsize \xmark} & {\scriptsize \xmark} \\
MSR-VTT~\cite{xu2016msr} & Open & 7,180 & M & {\scriptsize \xmark} & \cmark & {\scriptsize \xmark} & {\scriptsize \xmark} \\
UCF101~\cite{soomro2012ucf101} & Open & 13,320 & M & {\scriptsize \xmark} & \cmark & {\scriptsize \xmark} & {\scriptsize \xmark} \\
ActivityNetCap~\cite{krishna2017dense} & Activities & 20,000 & M & {\scriptsize \xmark} & \cmark & {\scriptsize \xmark} & {\scriptsize \xmark} \\
Shot2Story20k~\cite{han2023shot2story20k} & Open & 20,000 & M+S & {\scriptsize \xmark} & \cmark & {\scriptsize \xmark} & \cmark \\ \midrule
SumMe~\cite{gygli2014creating} & Events, holidays, sports & 25 & M & \cmark & {\scriptsize \xmark} & {\scriptsize \xmark} & {\scriptsize \xmark} \\
TVSum~\cite{song2015tvsum} & News, documentaries, vlogs & 50 & M & \cmark & {\scriptsize \xmark} & {\scriptsize \xmark} & {\scriptsize \xmark} \\
OVP~\cite{OVP} & Documentaries, lectures & 50 & M & \cmark & {\scriptsize \xmark} & {\scriptsize \xmark} & {\scriptsize \xmark} \\
VSUMM~\cite{de2011vsumm} & Cartoons, news, commercials, shows & 50 & M & \cmark & {\scriptsize \xmark} & {\scriptsize \xmark} & {\scriptsize \xmark} \\
EDUVSUM~\cite{ghauri2020eduvsum} & Letures & 98 & M & \cmark & {\scriptsize \xmark} & {\scriptsize \xmark} & {\scriptsize \xmark} \\
LoL~\cite{lol} & Matches of League of Legends & 218 & M & \cmark & {\scriptsize \xmark} & {\scriptsize \xmark} & {\scriptsize \xmark} \\
Ads-1K~\cite{tang2022multi} & Commercials & 1,041 & M+S & \cmark & {\scriptsize \xmark} & {\scriptsize \xmark} & {\scriptsize \xmark} \\
VideoXum~\cite{lin2023videoxum} & Activities & 14,001 & M & \cmark & \cmark & \cmark & {\scriptsize \xmark} \\
\rowcolor[HTML]{ECF4FF} \textbf{Instruct-V2Xum} & Open & \textbf{30,000} & M+S & \cmark & \cmark & \cmark & \cmark \\ \bottomrule
\end{tabular}%
}
\caption{Comparison with existing video-to-video summarization and video-to-text summarization datasets. ``V2V'', ``V2T'', and ``V2VT'' indicate support for video-to-video, video-to-text, or both tasks. ``Instruction'' denotes whether the dataset supports video-text instruction tuning. M and S stand for manual and model synthesized, respectively.}
\label{tab:dataset_compare}
\vspace{-3mm}
\end{table*}

%% file: sec/3_benchmark.tex
\section{The Instruct-V2Xum Dataset}

\subsection{Data Curation}
\subsubsection{Frame Captioning and Extractive Summarization.} 
The source videos are sampled from InternVid \cite{wang2023internvid}. We first filter the raw videos according to their duration and the aesthetic scores. The filtered data is then used to generate both video and text summaries. Specifically, we extract video frames at a rate of 1 FPS and then convert these frames into detailed textual descriptions using LLaVA-1.5-7B. After obtaining the textualized video frames, we utilize GPT-4V to perform extractive document summarization. Finally, the extracted summaries are converted into coherent video and text summaries.\\

\begin{figure*}[t]
    \centering
    \includegraphics[width=0.85\linewidth]{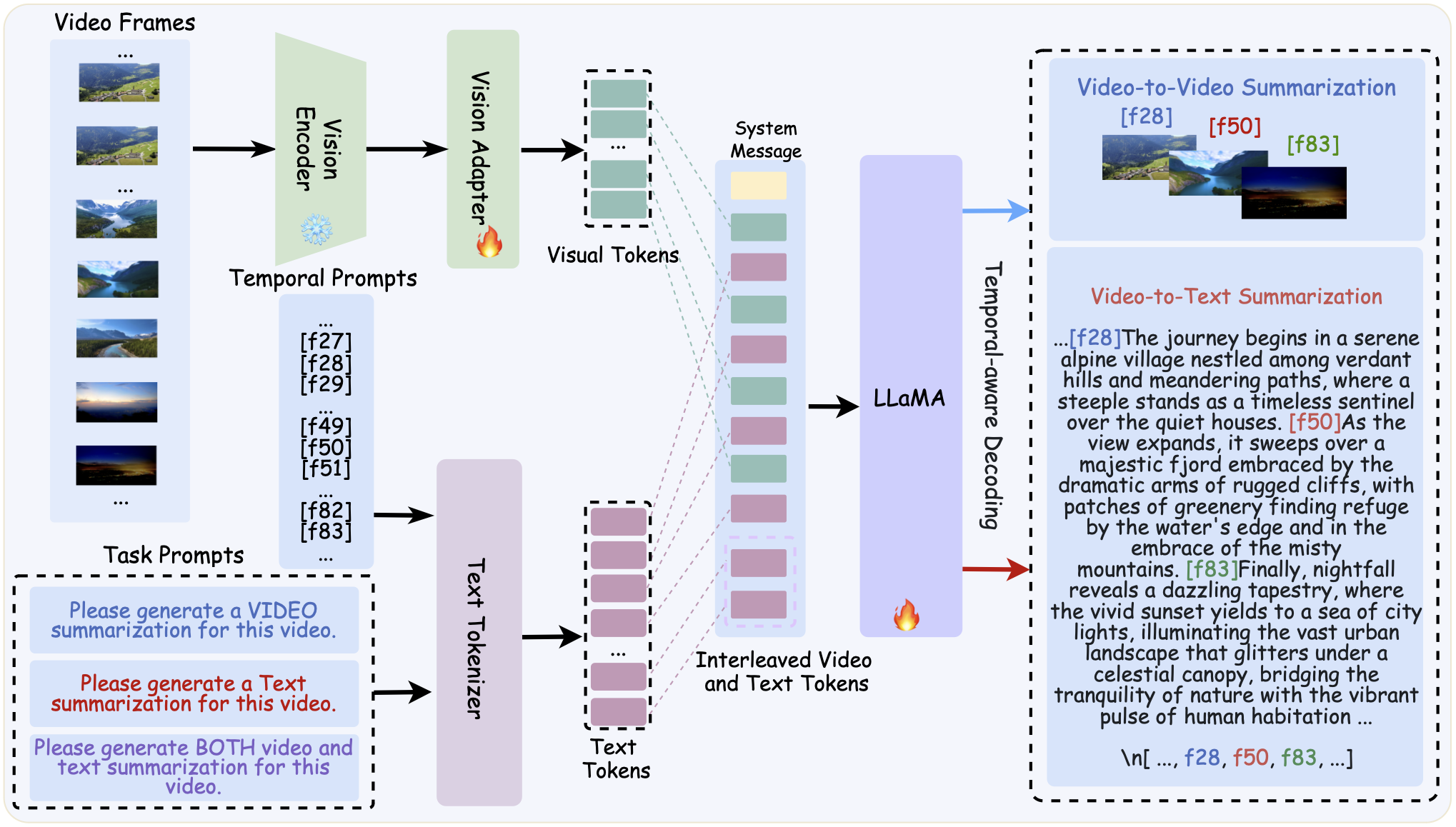}
    \vspace{-2mm}
    \caption{The architecture of the proposed V2Xum-LLaMA.}
    \label{fig:model}
\end{figure*}

\subsubsection{Text Summarization Refinement.} To further reduce the redundancy of the text summaries, we utilize BERT score \cite{zhang2019bertscore} to filter out the frame text representations that are similar to other summary frames. Here, we set the threshold to 0.93. The filtered video frame captions' indexes serve as the video summaries for the source videos. Then, we employ GPT-4 to further compress and rewrite the video summaries to be shorter and more grammar-fluent.
\subsubsection{Human Verification.}
To enhance the quality of the collected data, we employed human annotators to filter the GPT-4-generated data, resulting in a final set of 30k data points. We also show examples of human-filtered data in the Appendix.

\subsection{Quantity Analysis}
We provide the statistical results for our collected data.
Figure 6 in the appendix illustrates the distribution of the duration of source videos and the summaries length for both video and text in Instruct-V2Xum. We also provide more statistical results in the appendix. The average duration of the source videos is 183 seconds. The average text summary length is 239 tokens, and the average video summary length is 30 frames. The average compression ratio is 16.39\%.

\subsection{Quality Analysis}
We analyze the quality of the text summaries from the perspective of grammar fluency and commonsense plausibility. For the grammar-fluency evaluation, we leverage a grammar-check model \cite{morris2020textattack} that assigns high scores to correct texts grammatically. For the commonsense plausibility evaluation, we utilize
Vera \cite{liu2023vera}, a plausibility estimation model, to identify the nonsensical bias. The results of these evaluations are presented in the appendix. In summary, our human-filtered data demonstrates reasonable scores in both grammar and Vera evaluations.

%% file: sec/4_method.tex
\section{Methodology}

\subsection{Overview}
This section introduces our unified cross-modal video summarization framework, V2Xum-LLaMA, which employs the interleaved video frames along with temporal and task prompts as input and converts videos into multimodal summaries. \Cref{fig:model} depicts the architecture of our framework, where video frames are first encoded into visual representations, then combined with temporal and task prompts to obtain the interleaved input tokens for the LLaMA-2 decoder. Using the LLaMA decoder, our system can generate both video-to-video and video-to-text summaries. In addition, we design a temporal-aware decoding training strategy to enhance the alignment of text and video summaries. To sum up, our approach enables modality control, producing either video, text summaries, or both, and enhances the model’s understanding of the video timeline for more relevant and accurate outputs.
\subsection{Interleaved Video and Temporal Prompt Encoding}
The temporal prompt mechanism binds visual tokens to their corresponding frame-level timestamps, injecting positional information for each frame. To this end, we first encode each video frame $f_i$ using pretrained CLIP~\cite{clip} encoder $E_v$. The frames' encoding $v_i=E_v(f_i)$ are involved in the visual tokens sequence $V=\{v_1,v_2,...,v_L\}$, where $L$ is the number of sampled frames.

We then bind temporal prompts with each visual token. The temporal prompts are tokenized zero-padded numbers in natural language format like ``[f00]'', ``[f06]'', ``[f12]'', ``[f99]'', etc., indicated by $T=\{t_1, t_2, ..., t_L\}$. They are inserted into the visual token sequence $V$ to form a new sequence with interleaved visual tokens and temporal prompts:
\begin{equation}
    S=\{t_1, v_1, t_2, v_2, ..., t_L, v_L\}
\end{equation}
Compared to the original visual token sequence, the temporal prompted sequence can better capture the relations between the timestamps and the visual semantics. Then the sequence $S$ is projected into the word embedding space by the vision adapter, which is represented as $\overline{S}$.

\subsection{Temporal-Aware Decoding}
We use LLaMA-2~\cite{touvron2023llama2} as the decoder to generate V2V summarization $A^v$, V2T summarization $A^t$, or V2VT summarization $A^b$. The $I^v$, $I^t$, and $I^{b}$ represent the task instructions for V2V, V2T, and V2VT summarization. With the instruction $I$ and temporal prompted sequence $\overline{S}$, the output is given by the temporal-aware decoding:
\begin{equation}
    A^{x}=LLM(\overline{S}, I^{x}),~x\in\{v, t, b\}
\end{equation}
The V2V summarization $A^v$ is defined as a sequence of temporal tokens sequence, i.e., $A^v=\{v_i\}^M_{i=1}$, which is a subset of temporal prompts sequence $T$ and $M\le L$. The V2T summarization is a summarized caption in a natural language format. In our implementation shown as \Cref{fig:model}, the V2T summarization also contains frame referring temporal tokens: 
\begin{equation}
A^t=\{..., w_{i-1}, t_j, w_{i+1}, w_{i+2}...\}
\end{equation}
where the temporal token $t_j$ is bind with the words or sentences that consist of the summarization of visual content and can be further extracted from $A^t$ to get the V2V summarization $A^v$. This is called temporal-aware decoding. An example is shown in \cref{fig:model}. When decoding, the temporal tokens can be decoded together with the text summaries and represent the temporal position where the described visual content occurred in the input video.
\subsection{Task-Controllable Video Summarization Training}
As mentioned before, the types of summarization can be controlled by the task instructions $I^x$. Specifically, we use task prompts like ``Please generate a BOTH/VIDEO/TEXT summarization for this video.'' to instruct models to generate the corresponding video summaries. We then train the model end-to-end using negative log-likelihood loss:
\begin{equation}
    \mathcal{L} =- \sum_{\mathcal{D}} \sum_{i=1}^N \log p (A^{x}_i\mid \overline{S}, A^{x}_{\leq i-1}).
\end{equation}
where $\mathcal{D}$ denotes the training samples in the dataset, and $N$ is the length of the generated video and text summaries. During the training, we freeze the parameters of the vision encoder and update the vision adapter, and the language decoder to train the model to generate video and text summaries.

%% file: sec/5_experiments.tex
\section{Experiments}
\label{sec:expe}

\subsection{Baseline Models}
We benchmarked our V2Xum-LLaMA model, both 7B and 13B versions, against various models on V2V, V2T, and V2VT summarization tasks using the VideoXum dataset. Baseline models include LLM-based approaches such as Frozen-BLIP~\cite{li2023blip}, VSUM-BLIP~\cite{lin2023videoxum}, TSUM-BLIP~\cite{lin2023videoxum}, and VTSUM-BLIP~\cite{lin2023videoxum}. We compare with task-specific-head-free (TSH-Free) models like DENSE~\cite{krishna2017dense}, DVC-D-A~\cite{li2018jointly}, Bi-LSTM+TempoAttn~\cite{zhou2018end}, Masked Transformer~\cite{zhou2018end}, and Support-Set~\cite{patrick2020support}, which do not rely on regression-based timestamp prediction with extra task-specific heads. Additionally, on the classical TVSum~\cite{song2015tvsum} and SumMe~\cite{gygli2014creating} datasets, we compare our 7B version V2Xum-LLaMA with the following V2V summarization methods: dppLSTM~\cite{zhang2016video}, DSN~\cite{zhou2018dsn}, Sumgraph~\cite{park2020sumgraph}, CLIP-it~\cite{narasimhan2021clipit}, TL;DW~\cite{narasimhan2022tl}, iPTNet~\cite{iptnet}, A2Summ~\cite{a2summ}, Standard ranker~\cite{standardranker}, and VSUM-BLIP~\cite{lin2023videoxum}. We also evaluated Vid2Seq~\cite{yang2023vid2seq} models, initially pre-trained on the HowTo100M~\cite{miech2019howto100m}, VidChapter-7M~\cite{yang2024vidchapters}, YouCook2~\cite{youcook2}, and ViTT~\cite{huang2020vitt} datasets, and fine-tuned on VideoXum for fair comparison.
\input{table/model_compare}
\input{table/v2xum_eval}

\subsection{Evaluation Metrics}
We introduce new CLIP-based evaluation metrics for V2V and V2VT summarization evaluation. While the F1 score is a common metric for V2V summarization tasks, the process of video summarization annotation is highly subjective, leading to considerable variance among human annotators \cite{otani2019rethinking}. The traditional F1 score, which compares predicted video frames directly with the ground truth, fails to account for semantically similar frames that are close in time but not exactly matching, thus potentially undervaluing accurate summaries. To mitigate this, we introduce the $F_{CLIP}$ metric for V2V summarization evaluations, which is designed to recognize and reward semantic similarities between predicted and ground truth frames even when they are not identical. And we also propose the $Cross-F_{CLIP}$ metric for the V2VT summarization tasks. Unlike the VT-CLIPScore metric used in VideoXum—which calculates the average cross-modal CLIP score as an indicator of semantic alignment between predicted video and text summaries—our $Cross-F_{CLIP}$ calculates the average $F_{CLIP}$ scores between the predicted video summaries and the ground truth text summaries, as well as between the predicted text summaries and the ground truth video summaries. This approach aims to provide a more nuanced evaluation of summarization tasks by acknowledging the importance of semantic content alignment across modalities. For a reference video summary $v$ and predicted video summary $\senthyp$, the recall, precision, and F1 scores are:
\begin{align}
&\methodr (v,\senthyp) =\frac{1}{|v|} \sum_{v_i \in v}\max_{\senthyp_j \in \senthyp{}} \vref_i^\top \vhyp_{j}\\
&\methodp (v,\senthyp) = \frac{1}{|\senthyp|}  \sum_{\senthyp_j\in \senthyp}   \max_{v_i\in v}  \vref_{i}^\top \vhyp_{j}\\
&\methodf (v,\senthyp) = 2\frac{\methodp \cdot \methodr }{ \methodp + \methodr}
\end{align}
For the $Cross-F_{CLIP}$, given a reference video and text summary $v$ and $t$ and the predicted video summary $\senthyp$ and text summary $\txthyp$:
\begin{align}
Cross-F_{CLIP}(v,\senthyp,t,\txthyp)=\frac{\methodf(v,\txthyp)+\methodf(\senthyp,t)}{2}
\end{align}
Given that the cosine similarity values range from -1 to 1, we adjust the similarity scores by applying the operation $\max(\cos(\vref,\vhyp), 0)$. This ensures that only non-negative similarity scores are considered in our analysis.

\subsection{Implementation Details}
We use CLIP ViT-L/14@336 as the vision encoder and Vicuna-v1.5-7B/13B as the text decoder. The vision adapter is pre-trained with image-text pairs from LCS-558K~\cite{llava} dataset. Each tokenized temporal prompt takes four tokens. To facilitate model training, we normalize the video length into 100 via downsampling. For each setting, we train the V2Xum-LLaMA for 5 epochs with a learning rate of 1e-4 on 8$\times$NVIDIA A100 GPUs.

\subsection{Quantitative Results}
To verify the effectiveness of our method, we conducted experiments on various video summarization benchmarks, including cross-modal and V2V summarization tasks. For cross-modal video summarization, adopt the VideoXum dataset \cite{lin2023videoxum} and our proposed V2Xum dataset. For V2V summarization, we used the TVSum \cite{song2015tvsum} and SumMe \cite{gygli2014creating} benchmarks.

\subsubsection{Cross-Modal Video Summarization.}
We use the VideoXum dataset to evaluate our model's capability to cross-model video summarization. The experimental results are shown in \Cref{tab:model_compare}. It can be summarized that our proposed method outperforms all the baseline models. Specifically, in V2V summarization, V2Xum-LLaMA achieves a \textbf{8.1\%} higher F1-Score than VTSUM-BLIP, alongside significant improvements in both Spearman and Kendall correlation metrics. In addition, V2Xum-LLaMA-13B achieves higher evaluation scores on the V2V summarization task than V2Xum-LLaMA-7B. On the contrary, V2Xum-LLaMA-7B performs better on V2T summarization. We attribute this result to the increased complexity of the V2T summarization compared to V2V summarization. Additionally, the VideoXum training set comprises only 8,000 examples, a quantity insufficient for effectively training models with large language decoders. 

We also evaluate various models on our newly proposed Instruct-V2Xum dataset, as detailed in \Cref{tab:eval}. The results indicate that the models are well-adapted to the dataset, exhibiting sound performance. Moreover, V2Xum-LLaMA-13B outperforms V2Xum-LLaMA-7B in V2T summarization, which we believe is due to the larger language models benefiting from the increased volume of training data.

\subsubsection{Video-to-Video Summarization.}
We evaluate V2Xum-LLaMA on VideoXum, TVSum, SumMe, and Instruct-V2Xum datasets. The results are shown in Table \ref{tab:model_compare} and Table \ref{tab:tvsum_summe_compare}. It can be concluded that the unified video summarization using the language decoders in V2Xum-LLaMA can effectively perform traditional V2V summarization tasks. V2Xum-LLaMA outperforms all the previous methods that relied on task-specific regression heads for generating video summaries. This result indicates that LLMs with temporal prompts are capable of performing fine-grained video temporal understanding. The results presented in \Cref{tab:eval} affirm the validity of our proposed V2Xum dataset. It demonstrates that the model can properly fit the data.
\input{table/tvsum_summe_compare}
\input{table/ablation}
\subsection{Ablation Study}
To better evaluate the effectiveness of our proposed V2Xum-LLaMA framework and to underscore the importance of augmenting the training dataset, we conduct an ablation study on V2V and V2VT summarization tasks, detailed in Table \ref{tab:ablation}. First, we explore the performance enhancements attributed to the introduction of the temporal prompt mechanism. The results reveal that integrating the temporal prompt into the V2Xum-LLaMA framework yields superior results in both V2V and V2T summarization tasks, indicating that the temporal prompt benefits language models for fine-grained dense video temporal understanding. Second, we evaluated the simultaneous generation of video and text summaries, which enhanced performance in the V2V task. Third, we compared models trained with additional data (Instruct-V2Xum + VideoXum) to those without, showing that instruction data notably boosts performance, particularly in V2T summarization. Additionally, models with a pre-trained vision adapter outperformed those without, emphasizing the importance of pre-training for visual prompts.  Finally, we compared parameter-efficient fine-tuning (PEFT) to full parameter fine-tuning, with results favoring the latter in overall performance.

%% file: table/model_compare.tex
\begin{table*}[htpb]
\centering
\resizebox{\linewidth}{!}{%
\begin{tabular}{l|ccc|cccc|cccc|c}
\toprule
\multicolumn{1}{c|}{\multirow{2}{*}{\textbf{Method}}} & \multirow{2}{*}{\textbf{\begin{tabular}[c]{@{}c@{}}Cross-\\ Modal\end{tabular}}} & \multirow{2}{*}{\textbf{\begin{tabular}[c]{@{}c@{}}LLM-\\ Based\end{tabular}}} & \multirow{2}{*}{\textbf{\begin{tabular}[c]{@{}c@{}}TSH-\\ Free\end{tabular}}} & \multicolumn{4}{c|}{\textbf{V2T}} & \multicolumn{4}{c|}{\textbf{V2V}} & \textbf{V2VT} \\ \cline{5-13} 
\multicolumn{1}{c|}{} &  &  &  & B-4 & M & R-L & C & F1 & Spearman & Kendall & F$_{CLIP}$ & Cross-F$_{CLIP}$ \\ \midrule\midrule
DENSE~\cite{krishna2017dense} & {\scriptsize \xmark} & {\scriptsize \xmark} & \cmark & 1.6 & 8.9 & - & - & - & - & - & - & - \\
DVC-D-A~\cite{li2018jointly} & {\scriptsize \xmark} & {\scriptsize \xmark} & \cmark & 1.7 & 9.3 & - & - & - & - & - & - & - \\
Bi-LSTM+TempoAttn~\cite{zhou2018end} & {\scriptsize \xmark} & {\scriptsize \xmark} & \cmark & 2.1 & 10.0 & - & - & - & - & - & - & - \\
Masked Transformer~\cite{zhou2018end} & {\scriptsize \xmark} & {\scriptsize \xmark} & \cmark & 2.8 & 11.1 & - & - & - & - & - & - & - \\
Support-Set~\cite{patrick2020support} & {\scriptsize \xmark} & {\scriptsize \xmark} & \cmark & 1.5 & 6.9 & 17.8 & 3.2 & - & - & - & - & - \\
Frozen-BLIP~\cite{li2023blip} & \cmark & \cmark & {\scriptsize \xmark} & 0.0 & 0.4 & 1.4 & 0.0 & 16.1 & 0.011 & 0.008 & - & - \\
Vid2Seq-HCY~\cite{yang2023vid2seq} & \cmark & \cmark & \cmark & 2.3 & 8.2 & 19.0 & 7.6 & 24.2 & - & - & 0.888 & 0.214 \\
Vid2Seq-HC~\cite{yang2023vid2seq} & \cmark & \cmark & \cmark & 2.7 & 8.5 & 19.8 & 8.4 & 24.5 & - & - & 0.892 & 0.217 \\
Vid2Seq-HCV~\cite{yang2023vid2seq} & \cmark & \cmark & \cmark & 2.7 & 8.4 & 19.8 & 8.3 & 25.1 & - & - & 0.899 & 0.200 \\
VSUM-BLIP~\cite{lin2023videoxum} & {\scriptsize \xmark} & \cmark & {\scriptsize \xmark} & - & - & - & - & 21.7 & 0.207 & 0.131 & - & - \\
TSUM-BLIP~\cite{lin2023videoxum} & {\scriptsize \xmark} & \cmark & {\scriptsize \xmark} & 5.6 & 11.8 & 24.9 & 20.9 & - & - & - & - & - \\
VTSUM-BLIP~\cite{lin2023videoxum} & \cmark & \cmark & {\scriptsize \xmark} & \textbf{5.8} & 12.2 & 25.1 & 23.1 & 23.5 & 0.258 & 0.196 & 0.894 & 0.247 \\
\rowcolor[HTML]{ECF4FF}\textbf{V2Xum-LLaMA-7B (ours)} & \cmark & \cmark & \cmark & \textbf{5.8} & \textbf{12.3} & \textbf{26.3} & \textbf{26.9} & 29.0 & \textbf{0.298} & \textbf{0.204} & 0.931 & \textbf{0.253} \\
\rowcolor[HTML]{ECF4FF}\textbf{V2Xum-LLaMA-13B (ours)} & \cmark & \cmark & \cmark & 5.7 & \textbf{12.3} & 26.2 & 25.3 & \textbf{31.6} & 0.276 & 0.200 & \textbf{0.957} & 0.251 \\ \midrule
Human & \cmark & - & - & 5.2 & 14.7 & 25.7 & 24.2 & 33.8 & 0.305 & 0.336 & 0.944 & 0.256 \\ \bottomrule
\end{tabular}%
}
\caption{Comparison Results on the VideoXum dataset. ``TSH-Free'' indicates the model is task-specific-head-free; ``B-4'' denotes BLEU-4; ``M'' denotes METEOR; ``R-L'' refers to ROUGE-L metric; ``C'' represents CIDEr. For all metrics, higher scores indicate better performance.}
\label{tab:model_compare}
\vspace{-4mm}
\end{table*}

%% file: table/v2xum_eval.tex
\begin{table}[htbp]
\centering
\resizebox{\columnwidth}{!}{%
\begin{tabular}{l|cccc|cc|c}
\toprule
\multicolumn{1}{c|}{\multirow{2}{*}{\textbf{Method}}} & \multicolumn{4}{c|}{\textbf{V2T}} & \multicolumn{2}{c|}{\textbf{V2V}} & \textbf{V2TV} \\ \cline{2-8} 
\multicolumn{1}{c|}{} & B-4 & M & R-L & C & F1 & F$_{CLIP}$ & Cross-F$_{CLIP}$ \\ \midrule\midrule
Vid2Seq-HC~\cite{yang2023vid2seq} & 3.8 & 6.1 & 22.6 & 0.4 & 23.0 & 80.5 & 16.1 \\
Vid2Seq-HCY~\cite{yang2023vid2seq} & 3.7 & 6.2 & 22.4 & 0.5 & 24.7 & 81.3 & 16.0 \\
Vid2Seq-HCV~\cite{yang2023vid2seq} & 3.6 & 6.2 & 22.5 & 0.4 & 25.1 & 81.5 & 16.3 \\
\rowcolor[HTML]{ECF4FF}\textbf{V2Xum-LLaMA-7B} & \textbf{6.8} & \textbf{15.8} & 26.9 & \textbf{0.9} & \textbf{31.7} & \textbf{95.5} & \textbf{23.1} \\
\rowcolor[HTML]{ECF4FF}\textbf{V2Xum-LLaMA-13B} & 6.7 & \textbf{15.8} & \textbf{27.0} & 0.8 & 31.3 & 95.3 & 23.0 \\ \bottomrule
\end{tabular}%
}
\caption{Comparison results on the Instruct-V2Xum test set.}
\label{tab:eval}
\vspace{-4mm}
\end{table}

%% file: table/tvsum_summe_compare.tex
\begin{table}[t!]
\centering
\resizebox{\columnwidth}{!}{%
\begin{tabular}{l|cc|cc}
\toprule
\multicolumn{1}{c|}{\multirow{2}{*}{\textbf{Method}}} & \multicolumn{2}{c|}{\textbf{TVSum}} & \multicolumn{2}{c}{\textbf{SumMe}} \\ \cline{2-5} 
\multicolumn{1}{c|}{} & Spearman & Kendall & Spearman & Kendall \\ \midrule\midrule
dppLSTM~\cite{zhang2016video} & 0.055 & 0.042 & - & - \\
DSN~\cite{zhou2018dsn} & 0.020 & 0.026 & - & - \\
Sumgraph~\cite{park2020sumgraph} & 0.138 & 0.094 & - & - \\
CLIP-it~\cite{narasimhan2021clipit} & 0.147 & 0.108 & 0.120 & 0.109 \\
TL;DW~\cite{narasimhan2022tl} & 0.167 & 0.143 & 0.128 & 0.111 \\
iPTNet~\cite{iptnet} & 0.174 & 0.148 & 0.131 & 0.114 \\
A2Summ~\cite{a2summ} & 0.178 & 0.150 & 0.143 & 0.121 \\
Standard ranker~\cite{standardranker} & 0.230 & 0.176 & 0.014 & 0.011 \\
VSUM-BLIP~\cite{lin2023videoxum} & 0.261 & 0.200 & 0.365 & 0.268 \\
\rowcolor[HTML]{ECF4FF}\textbf{V2Xum-LLaMA} & \textbf{0.293} & \textbf{0.222} & \textbf{0.378} & \textbf{0.296} \\ \bottomrule
\end{tabular}%
}
\caption{Comparison results on the TVSum and SumMe datasets.}
\label{tab:tvsum_summe_compare}
\vspace{-4mm}
\end{table}

%% file: table/ablation.tex
\begin{table*}[htbp]
\centering
\resizebox{\linewidth}{!}{%
\begin{tabular}{l|cccc|cccc|c}
\toprule
\multicolumn{1}{c|}{\multirow{2}{*}{\textbf{Method}}} & \multicolumn{4}{c|}{\textbf{V2T}} & \multicolumn{4}{c|}{\textbf{V2V}} & \textbf{V2VT} \\ \cline{2-10} 
\multicolumn{1}{c|}{} & BLEU-4 & METEOR & ROUGE-L & CIDEr & F1-Score & Spearman & Kendall & F$_{CLIP}$ & Cross-F$_{CLIP}$ \\ \midrule\midrule
\rowcolor[HTML]{ECF4FF}\textbf{V2Xum-LLaMA} & \textbf{5.8} & \textbf{12.3} & \textbf{26.3} & \textbf{26.9} & \textbf{29.0} & \textbf{0.298} & \textbf{0.204} & \textbf{0.931} & \textbf{0.253} \\
w/o simultaneous VT-Sum & 5.6 & 12.2 & 25.6 & 25.9 & 25.1 & 0.249 & 0.203 & 0.926 & 0.251 \\
w/o Instruct-V2Xum & 4.9 & 12.0 & 24.3 & 21.6 & 23.1 & 0.260 & 0.191 & 0.921 & 0.252 \\
w/o fully fine-tuning & 4.5 & 11.7 & 24.7 & 22.8 & 23.4 & 0.222 & 0.175 & 0.915 & 0.250 \\
w/o temporal prompts & 4.4 & 11.7 & 24.4 & 21.2 & 23.9 & 0.258 & 0.192 & 0.910 & 0.249 \\
w/o pretrained adapter & 3.1 & 11.1 & 21.9 & 9.5 & 3.7 & - & - & - & - \\ \bottomrule
\end{tabular}%
}
\caption{Ablation Study of our V2Xum-LLaMA (7B) on the VideoXum dataset.}
\label{tab:ablation}
\vspace{-4mm}
\end{table*}

%% file: sec/6_revisiting.tex
\section{Limitations of Current Video Summarization: Data, Methods, and Evaluation}
\label{sec:revisit}
In this section, we discuss the limitations of existing vision summarization tasks from the perspective of data, method, and evaluation. 
As mentioned before, current existing video summarization datasets, including V2V and V2VT video summarization datasets, contain few training examples and cannot support training large-scale deep neural networks. The TVSum dataset comprises merely 50 YouTube videos; the SumMe dataset includes only 25 personal videos sourced from YouTube, while the QFVS dataset provides 135 video-query training samples. Additionally, VideoXum, the first cross-modal video summarization dataset, provides only 8k training examples. Our experimental results reveal that it is insufficient for training 13B models for the V2T summarization task. This issue is one of the significant drawbacks of current existing datasets. To address this problem, we collect more videos from YouTube and generate the corresponding cross-modal summaries for the videos using GPT-4 and propose a large-scale cross-modal video summarization dataset. 

A conventional approach to V2V summarization involves training a regression head to assign an importance score to each frame, ranking frames based on these scores, and selecting the top K\% frames to evaluate performance using metrics like the F1 score or Kendall/Spearman correlation against the ground truth. However, as demand for cross-modal video summarization grows, this method is inadequate for multimodal video summarization. Recent studies have begun to explore the use of language models for generating temporal and spatial references in videos \cite{huang2023vtimellm,li2024lego,ren2023timechat}, demonstrating the viability of using language models to generate the video interval indexes. In addition, leveraging large language models' text decoders for both V2V and V2T summarization tasks is a logical step. Therefore, in this study, we investigate how to prompt large VLMs to understand the fine-grained video content along with temporal information, how to maximally leverage the powerful capability of content understanding and reasoning of large language models, and how to achieve task controllability via natural language instructions.
To that end, we first propose the temporal prompt mechanism, and then design the visual encoder that takes the interleaved video frames, temporal, and language as input. Instead of using the regression head to perform V2V summarization, we unify different video summarization tasks into one language decoder.

Evaluating video summaries poses a significant challenge, primarily because the criteria for quality are inherently subjective, vary across different viewers and even fluctuate over time. This subjectivity, coupled with the limited availability of evaluation videos and annotations, further exacerbates the ambiguity in assessing video summary quality \cite{otani2019rethinking}. The traditional F1 score, designed to directly compare predicted video frames with ground truth, ignores the nuances of semantically similar frames that, while temporally proximate, do not exactly match. This oversight can lead to the undervaluation of otherwise accurate summaries.  To address this, we design new CLIP-based F scores for evaluating V2V and V2VT summarization tasks. Unlike the traditional F1 score, which relies on exact matches for its precision and recall calculations, $F_{CLIP}$ evaluates the V2V summarization from the perspective of semantic similarity. This approach allows for a more comprehensive yet meaningful evaluation that recognizes the importance of semantic accuracy over mere frame-by-frame accuracy. To evaluate the alignment of the generated video and text summaries, VideoXum employs an approach that calculates the average CLIP score across video frames and text summaries. This method involves computing vectors for all frames and sentences, applying mean pooling to these vectors to represent the video and text summaries, and then calculating the cosine similarity between them. However, this approach evaluates the video and text summaries as unified entities and thus neglects the detailed alignment at the sentence level with specific video frames. In contrast, our proposed $Cross-F_{CLIP}$ adopts a greedy matching strategy to optimize the similarity score between individual video frames and corresponding sentences, ensuring a more granular and accurate alignment.

%% file: sec/7_conclusion.tex
\section{Conclusion}
In this study, we address the deficiencies in current video summarization datasets including the insufficient number of training examples and the insufficient evaluation of video summarization by building a new large-scale cross-model video summarization dataset Instruct-V2Xum and designing the improved video summarization evaluation metrics. We also propose V2Xum-LLM, a novel temporal prompt instruction tuning method that unifies the generation of various video summary modalities within the text decoder of VLMs, eliminating the need for task-specific heads. This approach supports interleaved long video and language input sequences and allows modality-controllable summary generation through language instructions. Experimental results demonstrate the effectiveness of our method.


%% file: sec/8_appendix.tex
\appendix
\section{Appendix}

\subsection{Importance Score Prediction with LLM Logits}
Since our V2Xum-LLaMA directly utilizes natural language to represent the temporal positions, it is hard to obtain the corresponding importance scores that are usually adopted to calculate some conventional video summarization evaluation metrics, such as Spearman's $\rho$ and Kendall's $\tau$. Therefore, we use the logits from LLM that decode the temporal positions (numbers) as the importance scores.

Specifically, we first find the indices of the positions that are numbers in the output of V2Xum-LLaMA. Then, the corresponding logits are processed by the Softmax function to obtain the probabilities of the tens place and ones place:
\begin{align*}
p_i(tens) &= p_i(x_t|x_{<t}) = \text{SoftMax}(\text{logit}_t), \\
p_i(ones) &= p_i(x_{t+1}|x_{<t+1}) = \text{SoftMax}(\text{logit}_{t+1}),
\end{align*}
where $t$ is the index in the output sequence that the ten places of the temporal tokens occur. Next, we approximate \( p(\text{tens}, \text{ones}) \) with the product of \( p(\text{tens}) \) and \( p(\text{ones}) \):
$$
p_i(tens, ones)\approx p_i(tens)\times p_i(ones).
$$
Finally, we compute the average of all the $p(tens, ones)$ generated from the prediction in natural language format and treat it as important scores for calculating evaluation metrics:
$$
imp\_score = \frac{1}{M}\sum_{i=i}^{M}p_i(tens, ones).
$$

\subsection{Quality Analysis}
\begin{figure}[H]
    \centering
    \includegraphics[width=\linewidth]{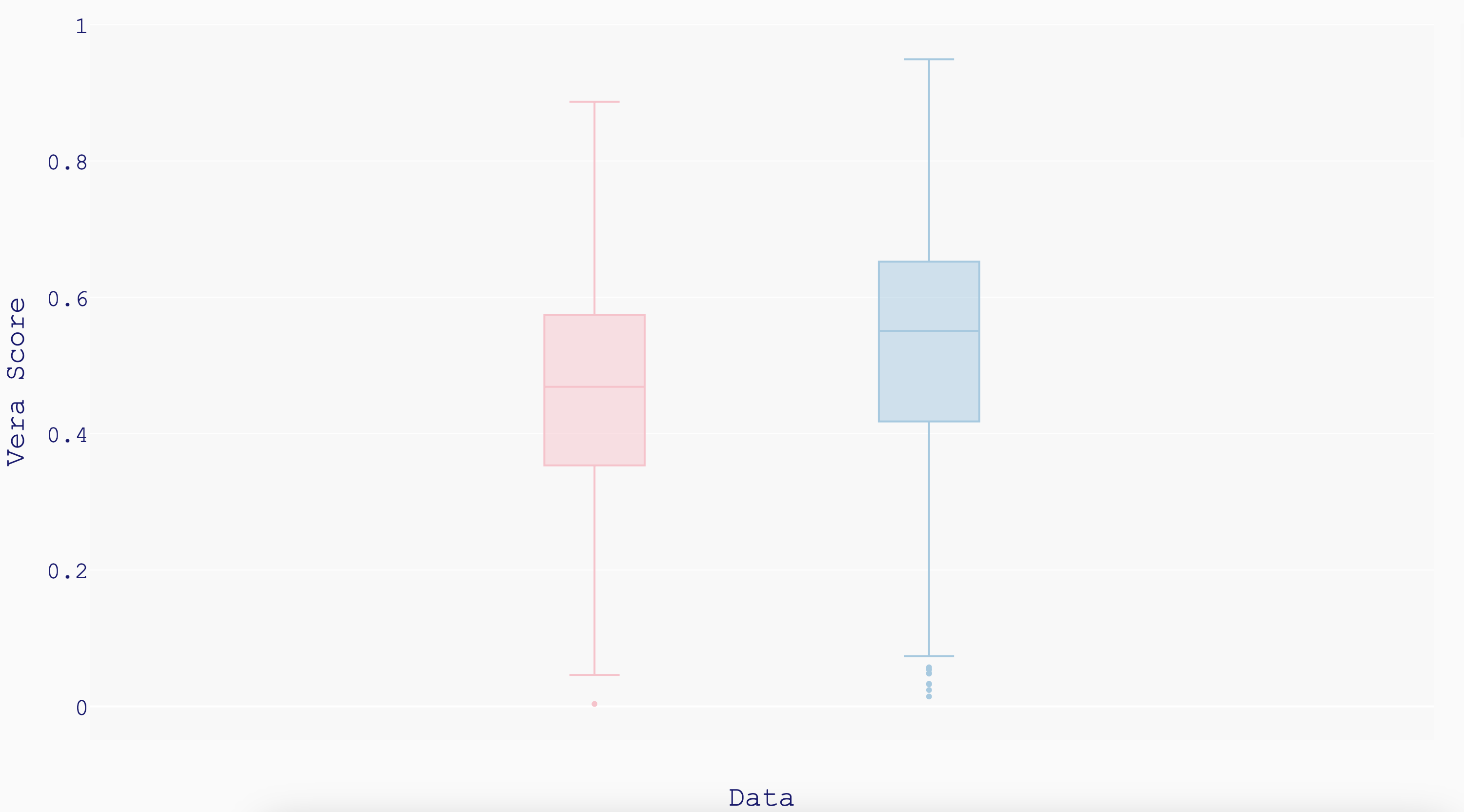}
    \caption{Comparison of the Vera scores for the dataset before (left) and after (right) refinement, with higher scores indicating better results.}
    \label{fig:vera}
\end{figure}

\begin{figure}[H]
    \centering
    \includegraphics[width=\linewidth]{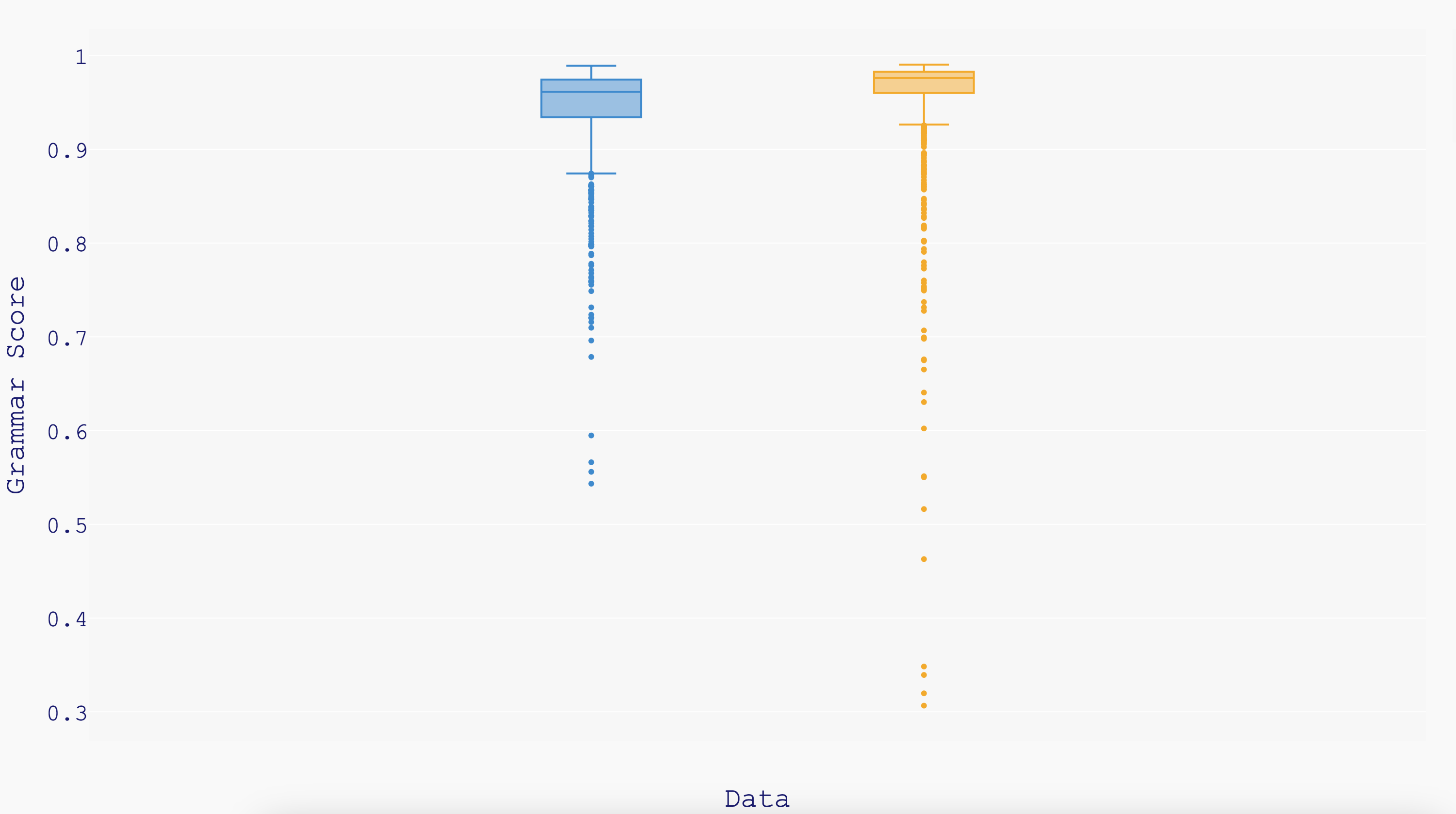}
    \caption{Comparison of the Grammar scores for the dataset before (left) and after (right) refinement, with higher scores indicating better results.}
    \label{fig:grammar}
\end{figure}

We analyze the quality of GPT-Synthesised data from the perspective of grammar fluency and commonsense plausibility. For the grammar-fluency evaluation, we leverage a grammar-check model \cite{morris2020textattack} that assigns high scores to correct texts grammatically. For the commonsense plausibility evaluation, we utilize
Vera \cite{liu2023vera}, a plausibility estimation model, to identify the nonsensical bias. The results are presented in Figure \ref{fig:vera} and Figure \ref{fig:grammar}. We can conclude that our dataset with the refinement process achieves a satisfactory score. 

\subsection{Example Data from Instruct-V2Xum}
\begin{figure*}
    \centering
    \includegraphics[width=0.82\linewidth]{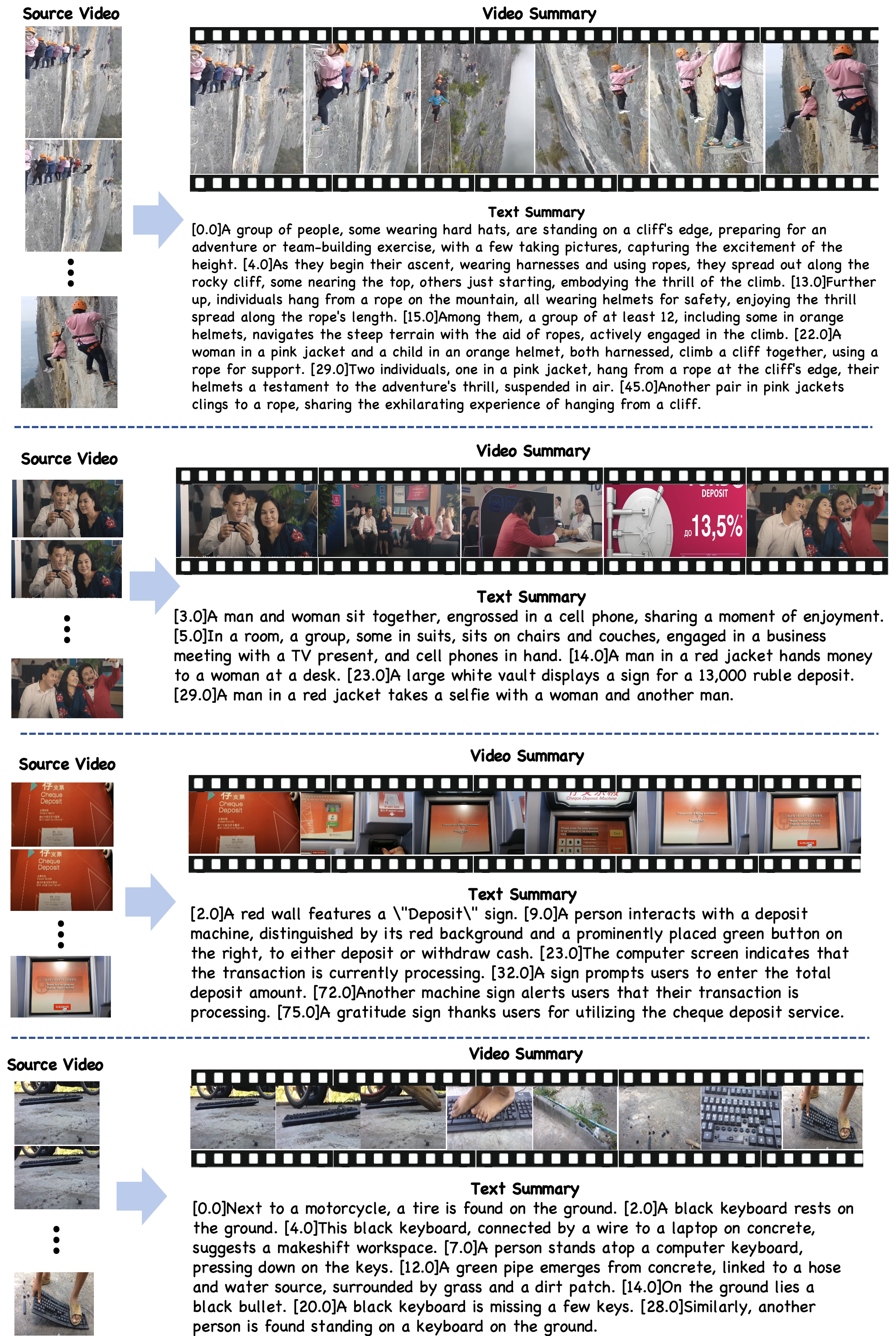}
    \caption{Some videos with video summaries and text summaries annotations from our Instruct-V2Xum dataset.}
    \label{fig:samples}
\end{figure*}

In this section, we show some data points sampled from the Instruct-V2Xum training set in Figure \ref{fig:samples}.

\begin{figure*}[htbp]
    \centering
    \includegraphics[width=0.9\linewidth]{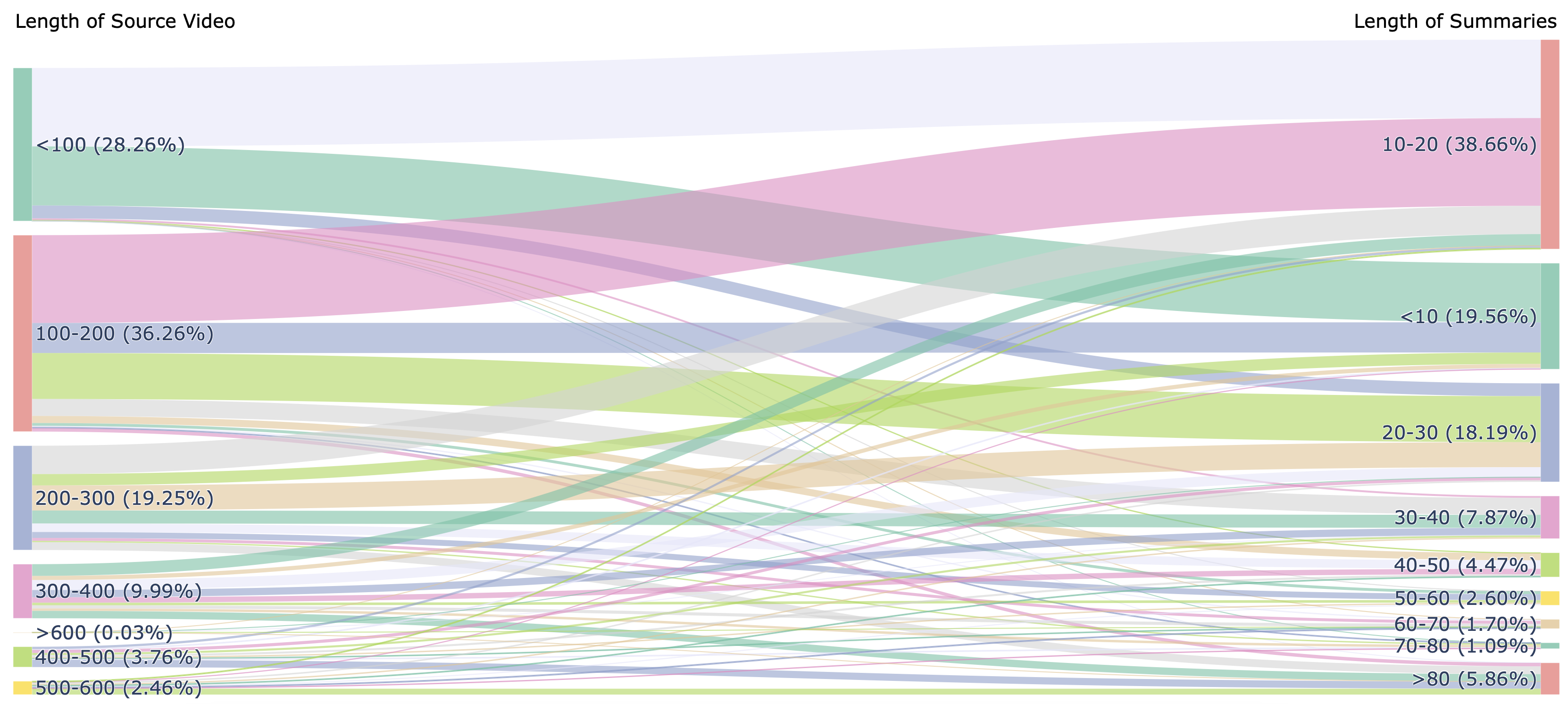}
    \caption{Distribution of the duration of source videos (in seconds) the summaries length for both video and text in Instruct-V2Xum. }
    \label{fig:quat1}
\end{figure*}

\subsection{V2Xum-LLaMA Prediction Visualization}
\begin{figure*}
    \centering
    \includegraphics[width=0.85\linewidth]{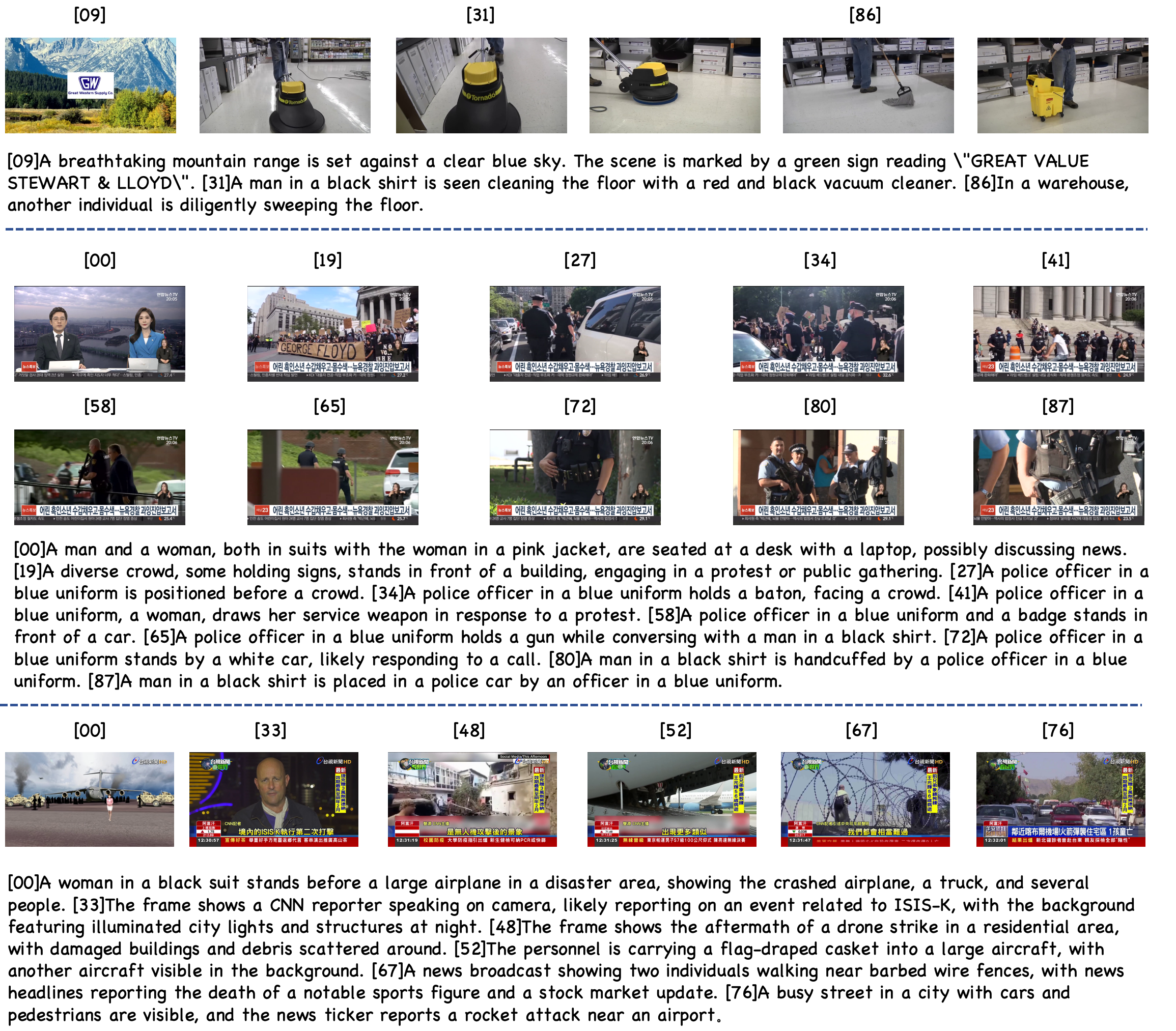}
    \caption{Some results predicted by V2Xum-LLaMA.}
    \label{fig:prediction}
\end{figure*}

In this section, we visualize some results predicted by V2Xum-LLaMA in Figure \ref{fig:prediction}.

